\documentclass[a4paper,11pt]{article}

\usepackage{authblk}
\usepackage{fancyhdr}
\usepackage[table,xcdraw]{xcolor}
\usepackage{makecell}
\usepackage{multirow}
\usepackage{array,arydshln}
\usepackage{paralist}    
\usepackage{booktabs}   
\usepackage{pifont}
\usepackage{amsmath}
\usepackage{amssymb}
\usepackage{mathtools}
\usepackage{microtype}
\usepackage{graphicx}
\usepackage{listings} 
\usepackage{color}
\usepackage[english]{babel}
\usepackage{txfonts}
\usepackage{url}
\usepackage{hyperref}
\usepackage{listings}  
\usepackage{afterpage}
\usepackage{verbatim}
\usepackage{subfig}
\usepackage{manyfoot}
\usepackage{caption}

\usepackage{enumitem}
\usepackage{textcomp}
\usepackage[normalem]{ulem}
\useunder{\uline}{\ul}{}

\newcommand{\extname}{Salp Swarm Optimization}
\newcommand{\ssa}{SSO}
\newcommand{\asso}{ASSO}

\providecommand{\keywords}[1]{\textbf{\textit{Keywords }} #1}

\begin{document}

\title{Salp Swarm Optimization: a Critical Review}

\author[1]{Mauro Castelli}
\author[2]{Luca Manzoni}
\author[3]{Luca Mariot}
\author[4,5]{Marco Nobile}
\author[6]{Andrea Tangherloni}

\affil[1]{{\scriptsize NOVA Information Management School (NOVA IMS), Lisboa, Portugal} 

	{\scriptsize \texttt{mcastelli@novaims.unl.pt}}}
	
\affil[2]{{\scriptsize Dipartimento di Matematica e Geoscienze, Universit\`a degli Studi di Trieste, Trieste, Italy}

	{\scriptsize \texttt{lmanzoni@units.it}}}
	
\affil[3]{{\scriptsize Cyber Security Research Group, Delft University of Technology, Delft, The Netherlands} 
	
	{\scriptsize \texttt{l.mariot@tudelft.nl}}}
\affil[4]{{\scriptsize Department of Industrial Engineering and Innovation Sciences, Eindhoven University of Technology, Eindhoven, The Netherlands}}

\affil[5]{{\scriptsize Bicocca Bioinformatics, Biostatistics and Bioengineering (B4) research center, Milano, Italy} 

{\scriptsize \texttt{m.s.nobile@tue.nl}}}

\affil[6]{{\scriptsize Department of Human and Social Sciences, University of Bergamo, Bergamo, Italy}

{\scriptsize \texttt{andrea.tangherloni@unibg.it}}}

\thispagestyle{fancy}

\maketitle

\begin{abstract}
In the crowded environment of bio-inspired population-based metaheuristics, the \extname{} (\ssa{}) algorithm recently appeared and immediately gained a lot of momentum.
Inspired by the peculiar spatial arrangement of salp colonies, which are displaced in long chains following a leader, this algorithm seems to provide an interesting optimization performance. 
However, the original work was characterized by some conceptual and mathematical flaws, which influenced all ensuing papers on the subject. In this manuscript, we perform a critical review of \ssa{}, highlighting all the issues present in the literature and their negative effects on the optimization process carried out by this algorithm. We also propose a mathematically correct version of \ssa{}, named Amended Salp Swarm Optimizer (\asso{}) that fixes all the discussed problems. We benchmarked the performance of \asso{} on a set of tailored experiments, showing that it is able to achieve better results than the original \ssa{}. Finally, we performed an extensive study aimed at understanding whether \ssa{} and its variants provide advantages compared to other metaheuristics. The experimental results, where \ssa{} cannot outperform simple well-known metaheuristics, suggest that the scientific community can safely abandon \ssa{}.
\end{abstract}


\thispagestyle{fancy}

\section{Introduction}
Most of the real-world problems can be formulated in terms of optimization problems, that is, they can be solved by finding the values that lead to the maximization, or minimization, of a target fitness function.
However, an analytic solution for such function is seldom available; moreover, these problems are often high-dimensional and are characterized by several local optima, where simple optimization algorithms easily get stuck.
In this context, bio-inspired population-based global optimization metaheuristics often proved their effectiveness in the identification of optimal solutions (i.e., global optima). 

A very large number of metaheuristics have been proposed in the latter years, taking inspiration among others from: Darwinian processes such as Differential Evolution \cite{price2006differential}, Evolution Strategies \cite{hansen2001completely}, Genetic Algorithms \cite{holland1992}, and all related variants \cite{viktorin2020dish}; the emergent collective behavior of groups of animals, including Particle Swarm Optimization \cite{poli2007particle,eberhart1995new}, Artificial Bee Colony \cite{karaboga2014comprehensive}, Ant Colony Optimization \cite{dorigo2006ant}, Bat Swarm Optimization \cite{yang2010new}, Firefly Algorithm \cite{yang2013firefly}, and all related variants \cite{nobile2018fuzzy}); finally, natural dynamic phenomena such as Gravitational Search \cite{rashedi2009gsa} and Artificial Immune Systems \cite{coello2005solving}.

Recently, a metaheuristics named \extname{} (\ssa{}), which is inspired by salp colonies, was proposed in~\cite{MIRJALILI2017163}. 
Salps are barrel-shaped animals living in swarms that are organized in long chains, actively looking for phytoplankton. 
\ssa{} leverages the metaphor of this peculiar spatial arrangement by translating it into an optimization algorithm.
Specifically, the individuals composing a population are constrained to follow the leader salp, while the latter is performing the actual exploration for food sources. 
In this metaphor, the leader explores the search space for optimal regions---with respect to the fitness function---while the followers exploit the area surrounding the leader.

\ssa{} quickly became popular as a real-valued~\cite{abualigah2019salp}, discrete~\cite{FARIS201843,ALJARAH2018964}, and even multi-objective  optimization algorithm \cite{kansal2020emended}.
This popularity apparently stems both from its performance and peculiar functioning paradigm, making the method to stand out among a plethora of seemingly novel algorithms~\cite{dorigo2020grey}.

In this paper, we show that the original algorithm is characterized by some conceptual and implementation flaws, hampering the performance of \ssa{} and all derived algorithms under some specific circumstances.
In particular, we will show that the most remarkable issue is the fact that \ssa{}, in its original formulation, is not shift-invariant. 
This raises a problem if the lower bound of the search space is too far from the origin.
In this paper, we first propose an Amended Salp Swarm Optimizer (\asso{}), which represents a correct reformulation of the \ssa{} algorithm as was probably intended by its creators.	Still, our objective is not to promote the use of \asso{} in the scientific community. Indeed, while \asso{} fixes the mathematical flaws of \ssa{}, it is still not grounded on any theoretical basis. Thus, \asso{} would be one of the several metaheuristics based on some natural metaphor and built by combining search operators that do not rely on any theoretical property~\cite{piotrowski2018some}. For this reason, we subsequently assess the performance of \asso{} and \ssa{} against simple and well-known metaheuristics on a set of benchmark functions. This analysis aims at understanding whether the scientific community could abandon the use of \ssa{} and \asso{}, or if these algorithms provide some advantages with respect to simpler metaheuristics widely established in this research field. As a side note, \ssa{} changed name several times in the literature: indeed, we can find it as Salp Swarm Algorithm \cite{sayed2018novel}, Salp Swarm Optimizer \cite{el2018extracting,8970592}, Salp Swarm Optimization \cite{cheng2020new}, Salp Optimization Algorithm \cite{tian2020improved}, and so forth.
Incidentally, in the highlights of the original paper \cite{MIRJALILI2017163}, the authors state that the official name of the algorithm is Salp Swarm Optimizer, although the acronym SSA is used in the rest of the article. In this paper, we will use the name \extname{} and the acronym \ssa{} to prevent confusion. We wish to emphasize that this work aims at discussing the limitations and issues of the original \ssa{} algorithm, showing how even a simple Random Search (RS) algorithm can outperform the original incorrect \ssa{} algorithm, under specific circumstances.
Therefore, we think that our results and findings raise an alert on the existing \ssa{} literature. 

This paper is structured as follows. Section~\ref{sec:litreview} reviews the related works highlighting the flaws and limitations of some of the recently defined metaheuristics. Section~\ref{sec:review} provides a critical review of \ssa{}, in particular concerning the methodological issues with the updating rule of the position of the leader salp and the physically-inspired motivation for the updating rule of the positions of the follower salps. Section~\ref{sec:exp} presents an experimental evaluation of the improved version of \ssa{} that fixes some of the raised issues, and compares the result with the original version on a set of standard benchmark functions. Subsequently, we analyze the performance of \ssa{} and its variants on the CEC 2017 benchmark function suite against two commonly used metaheuristics, namely Covariance Matrix Adaptation Evolution Strategy (CMAES)~\cite{hansen1996adapting} and Differential Evolution (DE)~\cite{storn1997differential}. Finally, Section~\ref{sec:conc} summarises the main issues of the original \ssa{} algorithm, and concludes with a broader remark on the nature of the salp metaphor.

\section{Related work}
\label{sec:litreview}
Recent years have witnessed the definition of a significant number of metaheuristics inspired by some natural phenomenon~\cite{fong2017meta}.
The common idea behind the definition of these metaheuristics is to consider a natural process and, subsequently, to design the underlying metaheuristic by exploiting the natural metaphor observed.
After the publication of a given metaheuristic, it is also common to see a significant number of scientific papers that use the new metaheuristic to address complex real-world problems claiming its superior performance compared to existing metaheuristics. 
Luckily, a research strand started to criticize the definition of these metaheuristics, observing that, in most of the cases, their performance cannot be better than the commonly used evolutionary strategies~\cite{weyland2015critical}. 
Despite the lack of novelty that characterizes these metaheuristics (i.e., the change concerns only the underlying natural metaphor), a fundamental issue is that some of the results achieved through them are not reliable.

In this sense, one of the clearest examples is the paper by Weyland~\cite{weyland2015critical}, in which the author demonstrated that Harmony Search (HS)~\cite{lee2005new} cannot be used to successfully solve a sudoku, thus contradicting the results obtained by Geem~\cite{geem2007harmony}.
More precisely, Weyland first proved that HS is a special case of evolution strategies, thus highlighting the lack of novelty of the metaheuristic.
As a consequence, the performance of HS is always bounded by the performance that can be obtained by evolution strategies.
Finally, Weyland demonstrated that the results achieved in~\cite{geem2007harmony} are flawed both from the theoretical and the practical point of view, concluding that there is no reason for the existence of HS as a novel metaheuristic.
Despite the Weyland's work clearly demonstrated the lack of novelty of HS, the algorithm is still largely used nowadays.
Further, even though Weyland proved beyond any doubt that HS cannot perform better than evolutionary strategies, several papers still claim its presumed superior performance~\cite{ala2019comprehensive}.

Thus, it seems that practitioners are still deceived by metaheuristics whose novelty is based only on the use of some natural metaphors.
The truth is that HS, and other related metaheuristics, are simply using a different terminology with respect to the classic evolutionary strategies.
While the lack of novelty did not prevent metaheuristics based on natural metaphors to be published in well-renowned scientific venues, some scientific journals are seriously tackling the problem.
For instance, Marco Dorigo, the editor in chief of Swarm Intelligence, published an editorial note~\cite{dorigo2016swarm} stating that he observed a new trend consisting in ``taking a natural system/process and use it as a metaphor to generate an algorithm whose components have names taken from the natural system/process used as metaphor''.  
Dorigo also highlighted that ``this approach has become so common that there are now hundreds of so-called new algorithms that are submitted (and unfortunately often also published) to journals and conferences every year'', and concluded his editorial stating that ``it is difficult to understand what is new and what is the same as the old with just a new name, and whether the proposed algorithm is just a small incremental improvement of a known algorithm or a radically new idea''.

A similar analysis on this trend appeared in the work by Cruz \textit{et al.}~\cite{cruz2019critical}. There, the authors first highlighted the vast number of swarm intelligence algorithms developed by taking inspiration from the behavior of insects and other animals and phenomena.
Subsequently, they showed that most algorithms present common macro-processes among them, despite the fact that they are inspired by different metaphors.
In other words, the considered metaheuristics are characterized by common issues and features which happen at the individual level, promoting very similar emergent phenomena~\cite{cruz2019critical}.
Thus, it is difficult (if not impossible) to claim that such metaheuristics are really novel.

Focusing on specific metaheuristics, some contributions that analyze the behavior of a given algorithm started to appear~\cite{camacho2019intelligent,niu2019defect,villalon2021cuckoo}.
In~\cite{camacho2019intelligent}, Villal\'on \textit{et al.} thoroughly investigated the Intelligent Water Drops (IWD) algorithm~\cite{hosseini2007problem}, a metaheuristic proposed to address discrete optimization problems.
The authors demonstrated that the main steps of the IWD algorithm are special cases of Ant Colony Optimization (ACO)~\cite{dorigo2006ant}.
Thus, the performance of IWD cannot be better than the best ACO algorithm.
Moreover, the authors analyzed the metaphor used for the IWD definition, and from their analysis, the metaphor is based on ``unconvincing assumptions of river dynamics and soil erosion that lack a real scientific rationale''. 
Finally, they pointed out that the improvements proposed to the IWD algorithm are based on ideas and concepts already investigated in the literature many years before in the context of ACO.

Niu \textit{et al.}~\cite{niu2019defect} analyzed the Grey Wolf Optimization (GWO) algorithm~\cite{mirjalili2014grey}, and demonstrated that, despite its popularity, GWO is flawed.
In particular, GWO shows good performance for optimization problems whose optimal solution is $0$, while the same performance cannot be obtained if the optimal solution is shifted.
In particular, when GWO solves the same optimization function, the farther the function’s optimal solution is from $0$, the worse its performance is.
Interestingly, GWO was proposed by the same author of the SSO algorithm analyzed in this paper, and it presents some of the SSO's flaws.

Villal\'on \textit{et al.}~\cite{villalon2021cuckoo} analyzed the popular Cuckoo Search (CS) algorithm~\cite{yang2010engineering}, a metaheuristic introduced in 2009.
The authors analyzed CS from a theoretical standpoint, and showed that it is based on the same concepts as those proposed in the $(\mu + \lambda)$ evolution strategy proposed in 1981~\cite{schwefel1981numerical}.
Further, the authors evaluated the algorithm and the metaphor used for its definition based on four criteria (i.e., usefulness, novelty, dispensability, and sound motivation), and they concluded that CS does not comply with any of these criteria.
Finally, they pointed out that the original CS algorithm does not match the publicly available implementation of the algorithm provided by the authors of the algorithm.

This analysis is quite surprising, given the popularity of CS, and it highlights the need for a thorough investigation of the existing metaheuristics, with the goal of understanding which of them should be abandoned by the scientific community.
Indeed, the impressive number of metaheuristics published in the literature makes it difficult to determine whether they really contribute to the advancement of the field.
This problem was pointed out by Mart\'inez \textit{et al.}~\cite{garcia2017since}.
Specifically, the authors investigated whether the increasing number of publications is correlated with real progress in the field of heuristic-based optimization. 
To answer this research question, the authors compared five heuristics proposed in some of the most reputed journals in the area, and compared their performance to the winner of the IEEE Congress on Evolutionary Computation 2005.
The results showed that the considered methods could not achieve the result of the competition winner, which was published several years before.
Moreover, a comparison with the state-of-the-art algorithms is often missing, thus making it impossible to understand the real advantage provided by a new method.

In the same vein, Piotrowski and Napiorkowski~\cite{piotrowski2018some} highlighted the risk associated with the definition of new and increasingly complex optimization metaheuristics and the introduction of structural bias~\cite{kononova2015structural}.
In particular, the authors focused on two winners of the CEC 2016 competition, and they found out that each of them includes a procedure that introduces a structural bias by attracting the population towards the origin. As a final message, the authors highlighted that some metaheuristics have to be simplified because they contain operators that structurally bias their search, while other metaheuristics should be simplified (or abandoned) as they use unnecessary operators. 

\section{Critical review of SSO}
\label{sec:review}

This section outlines the design issues we found in the original definition and implementation of the \ssa{} algorithm.
More precisely, the following issues are discussed here and in Section~\ref{sec:exp}:
\begin{itemize}
	\item  the update rule of the main salp does not work correctly when one of the dimensions has a lower bound different from zero (Section~\ref{sec:update}). In the experimental part, we show that simply shifting a 2D sphere function from the origin makes \ssa{} perform worse than a simple RS (Section~\ref{sec:exp});
	\item the physical motivations for the updating rule of the follower salps are incorrectly derived from Newton's laws of motion (Section~\ref{sec:physical});
	\item there is a clear divergence between the algorithm described in~\cite{MIRJALILI2017163} and the available \ssa{} implementation.
	This makes it difficult, or even impossible, to compare results from different papers (Section~\ref{sec:implem});
	\item finally, we experimentally show that the original \ssa{} algorithm has a bias towards the origin.
	Since many of the considered benchmark problems have the optimum in the origin, the results are biased in favor of \ssa{} (Section~\ref{sec:exp}).
\end{itemize}

In what follows, we will refer to the equations that were introduced in the original \ssa{} paper by \cite{MIRJALILI2017163}.

\subsection{Leader Salp Updating Rule}
\label{sec:update}

We assume here that a chain of $N$ different salps moves within a bounded $D$-dimensional search space, aiming at identifying the optimal solution.

The first issue is related to the definition of the updating rule for the position of the leader salp.
In Equation~3.1 of the original paper~\cite{MIRJALILI2017163}, the update for the leader salp along the $j$-th dimension, with $j=1, \dots, D$, is given as follows:
\begin{align}
x_j^1 = 
\begin{cases}
F_j + c_1 ((ul_j - lb_j)c_2 + lb_j) & \text{if $c_3 \ge 0$} \\
F_j - c_1 ((ul_j - lb_j)c_2 + lb_j) & \text{if $c_3 < 0$}
\end{cases} \enspace ,
\label{eq:3.1}
\end{align}
where the upper and lower bounds of the search space in the $j$-th dimension are $ul_j$ and $lb_j$, respectively.
The value $F_j \in [lb_j, ul_j]$ is the position of the best solution found so far in the $j$-th dimension, that corresponds to the best food source.
On the other hand, $c_1$ decreases exponentially with the number of iterations according to the following rule:
\begin{align*}
c_1 = 2 e^{-(\frac{4\ell}{L})^2} \enspace ,
\end{align*}
where $\ell$ is the current iteration number while $L$ is the total number of iterations. 
Finally, both $c_2$ and $c_3$ are random numbers selected uniformly in the range $[0,1]$.

Here, we can find the first definitional issue, although it did not propagate in the original source code.
Indeed, $c_3 \in [0,1]$ implies that the second case of Equation~\eqref{eq:3.1} is never verified.
However, in the source code the threshold for $c_3$ is set to $0.5$ instead of $0$, which means that the two cases of the updating rule occur with equal probability. This kind of oversights is usually not a problem, but several successive papers \emph{do not correct} the issue (see, e.g., \cite{al2020optimization,alresheedi2019improved,sayed2018novel,ibrahim2019improved,yang2019novel,hussien2017swarming,tolba2018novel,yang2019hybrid,ekinci2018parameter}).
Further, if a new implementation following the specifications of~\cite{MIRJALILI2017163} is used, instead of the original one released by the authors, the results might not be comparable.
It is currently unknown how many papers on \ssa{} employ the same implementation of the original paper.

The main issue with the updating rule is however more significant, and it is still present in the source code and widespread across all existing \ssa{} variants (\cite{al2020optimization,alresheedi2019improved,sayed2018novel,ibrahim2019improved,yang2019novel,el2018extracting,ABBASSI2019362,aljarah2018asynchronous,hussien2017swarming,tolba2018novel,ahmed2018feature,yang2019hybrid,qais2019enhanced,ekinci2018parameter}).
Let us consider the variation with respect to $F_j$, which is:
\begin{align*}
c_1((ul_j - lb_j)c_2 + lb_j) = c_1 c_2 (ul_j - lb_j) + c_1 lb_j \enspace.
\end{align*}
Notice that while $c_1 c_2 (ul_j - lb_j)$ gives a value in $[0, ul_j - lb_j]$ (i.e., $F_j + c_1 c_2 (ul_j - lb_j)$ might remain inside the search space), the value $c_1 lb_j$ is also added and it can be arbitrarily large in absolute value. 
Although this is not the case in the experiments performed in the original paper, simply shifting the search space by a suitably large constant might significantly hamper the search process of \ssa{}.
The issue could also potentially affect research in applied disciplines where \ssa{} has been used (e.g., COVID-19 related research, as proposed by \cite{al2020optimization}).

As an example, let us suppose that the lower and the upper bounds are of the same order of magnitude for all the $D$ dimensions, and in particular that $lb_j= 10^k$ and $ub_j = 10^k + 1$, with $j=1, \dots, D$.
In other words, the search space is the hypercube $[10^k,10^k+1]^j$.
By using the updating rule given in Equation~\eqref{eq:3.1}, the position of the leader salp is updated as
\begin{equation}
\begin{aligned}
x_j^1 &= F_j \pm c_1 c_2 (ul_j - lb_j) + c_1 lb_j \\
&= F_j \pm c_1c_2 + c_1 10^k
\label{eq:ublb}
\end{aligned}\enspace .
\end{equation}
Since $c_1 c_2 \le 2$ (recall that $c_2 \in [0,1]$), the position update in Equation~\eqref{eq:ublb} is dominated by the $c_1 10^k$ term.
Consequently, this update will move the leader salp out of the admissible bounds for most of the values taken by $c_1$, forcing its position to be clipped on the borders of the search space most of the times.
This implies that \ssa{} is not invariant with respect to translations of the search space.
Indeed, given that $c_1 = 2 e^{-(\frac{4\ell}{L})^2}$, the salp remains inside the search space $[10^k,10^k+1]^j$ only if $c_1 10^k \le 1$, namely:
\begin{align*}
2 e^{-(\frac{4\ell}{L})^2} 10^k \le 1 \ \Rightarrow \ k \le \log_{10} \left ( \frac{1}{2}e^{-(\frac{4\ell}{L})^2} \right )\enspace .
\end{align*}
The effect is that, depending on the search space, the majority of the updates of the leader salp can force it on the boundary of the space (due to clipping), with only the later iterations (with $c_1$ small enough) resulting in the leaders salp moving without clipping. 
In particular $e^{-(\frac{4\ell}{L})^2}$ yields the smallest value when $\ell = L$, i.e., at the last iteration and in this case we obtain that
\begin{align*}
k \le \log_{10} \left ( \frac{1}{2}e^{-16} \right ) \approx 6.648 \enspace .
\end{align*}
Therefore, $k > 6.648$ will result in a search space where the position of the leader salp will \emph{always} be forced outside of the search space, or equivalently, the leader salp will continue to ``bounce'' on the boundaries of the search space.

Similar pathological examples can be found by tweaking the values of the upper and lower bounds.
This observation reveals that the initial value of $c_1$ must be carefully chosen with respect to the size and the shift of the search space. 
In other words, \ssa{} is also not invariant with respect to rescalings of the search space.

Another different issue is that for any dimension $j$, the quantity $c_2(ul_j - lb_j) + lb_j$ has an expected value of $\frac{ul_j + lb_j}{2}$. When the search space is centered in zero the expected value is then zero. As we will see in the experimental part, this gives an unfair advantage to problems where the search space is symmetric (with respect to $\textbf{0}$) and the global optimum is in $\textbf{0}$.

\subsection{Physically-inspired motivations}
\label{sec:physical}

The original paper claimed that the definition of the updating rule for the follower salps is based on the principles of classical mechanics (Newton's laws).
However, there are important issues concerning the formulation of this rule, as well as the correct use of Newton's laws of motion.
The equation for the update of the follower salps in \ssa{} is:
\begin{equation}
\label{eq:acc}
x_j^i = \frac 1 2 a t^2 + v_0 t \enspace ,
\end{equation}
where it can be assumed that $a=a_j^i$ and $v_0=v_j^i$ at $t=0$, for each dimension $j$ (with $j=1, \dots, D$) and follower salp $i$ (with $i=1, \dots, N$). 
In the original paper, the acceleration $a$ is calculated as: 
\begin{equation}
\label{eq:acc2}
a = \frac{v_\mathrm{final}}{v_0} \enspace ,
\end{equation}
which is incorrect, since the average acceleration in a time interval $\Delta t$ is:
\begin{equation*}
a = \frac{v(t+\Delta t) - v(t)}{\Delta t} \enspace .
\end{equation*}
Notice that the incorrect formula also gives a dimensionless quantity, instead of a length divided by a squared time. 
Notably, this issue is only sometimes corrected (with $a = v_{\mathrm{final}} - v_0$) in the \ssa{} literature.

In addition, the formula $v_{final} = \frac{x - x_0}{t}$, with $x$ and $x_0$ being the final and initial positions, respectively, and $t$ being the time interval, is correct for the \emph{average speed}, which is however not necessary in all the derivations of the paper. 
In fact, for computing the average acceleration, the \emph{instantaneous speed} must be used instead. 
Furthermore, the aim of the derivation is to obtain the final position of the salp, which we cannot use to compute the average speed.
Since the salps are initially still, the original definition of \ssa{} \cite{MIRJALILI2017163} explicitly uses $v_0 = 0$  which, when substituted in the Equations \ref{eq:acc} and \ref{eq:acc2}, gives an infinite acceleration that would lead the salps outside any boundary of the search space. 

Regardless of these issues, the authors eventually point out that $t$, in the equations, corresponds to the iteration number, so that $\Delta t = 1$ and the time term can be cancelled out from the equations.
They conclude that this step leads to the final formula:
\begin{equation}
\label{eq:3.4}
x_j^i = \frac{1}{2}(x_j^i + x_j^{i-1}) \enspace .
\end{equation}
Unfortunately, this equation cannot be derived from the previous ones.
In fact, the position of the next salp in the chain never appears before this point and it is not taken into account in any of the previous derivations.

A correct way to derive the previous formula would be the following.
Assume that the $i$-th salp is moving toward the current position of the $(i-1)$-th salp (in the $j$-th dimension) with starting speed of $0$, final speed of $1$ unit for time step, and constant acceleration of $(x_j^i(t) - x_j^{i-1}(t))$ units for time step squared.
Hence, the new position after one time unit can be computed as follows:
\begin{equation*}
\begin{aligned}
x_j^i(t+1) & = \frac{1}{2}(x_j^{i-1}(t) - x_j^i(t)) + x_j^i(t) \\
& = \frac{1}{2}(x_j^i(t) + x_j^{i-1}(t)).
\end{aligned} 
\end{equation*}
Notice that this is only one of the possible ways to \emph{correctly} derive the updating rule for the follower salps, but it is completely unrelated to the biological metaphor exploited by the authors.
This critique only concerns the physical motivations for the definition of the updating rules, not the updating rule itself.
However, the flawed explanation presented in the original paper is restated into multiple papers~\cite{ibrahim2019improved,yang2019novel,hussien2017swarming,tolba2018novel}, without any significant correction.

\subsection{Implementation Issues}
\label{sec:implem}
It is worth mentioning that both the MATLAB\textsuperscript{\textregistered} and Python implementations\footnote{The source code of both implementations is available at the following address: \texttt{\url{http://www.alimirjalili.com/SSA.html}}} do not correctly implement the pseudo-code and, unfortunately, do not follow the explanations provided by the authors in the original paper~\cite{MIRJALILI2017163}. 
As a matter of fact, in both implementations, the authors updated the first half of the population using Equation~\ref{eq:3.1} and the second half using Equation~\ref{eq:3.4}.
Considering that the salp chain is composed of $N$ different individuals, the first $\frac{N}{2}-1$ salps perform the exploration process for food sources attracted by the best food source found so far (updated by Equation~\ref{eq:3.1}). 
The $\frac{N}{2}$-th individual is the leader salp that drags the follower salps, which exploit the area surrounding the leader (updated by Equation~\ref{eq:3.4}). 
In what follows, we will refer to this implementation as \ssa{}-code. 

We modified the implementation of \ssa{}-code by removing the term $c_1 lb_j$ from Equation~\eqref{eq:3.1}, for each dimension $j$ (with $j=1, \dots, D$).
In such a way, we are able to avoid as much as possible the clipping step of the salp positions due to the wrong update, proposed in the original Equation (i.e., Equation~\ref{eq:3.1}), which sends the salps out of the admissible bounds.
We will refer to this version as ASSO.

\section{Experimental Evaluation}
\label{sec:exp}

As a first batch of tests, we compared the performance of a simple Random Search (RS), \ssa{}, \ssa{}-code, and \asso{} using $2$-$D$ standard benchmark functions (i.e., Ackley, Alpine, Rosenbrock, and Sphere).
Then, the RS, \ssa{}, \ssa{}-code, and \asso{} were compared against basic versions of CMAES \cite{hansen1996adapting} and DE \cite{storn1997differential}.
We used the standard version of CMAES and DE implemented in the Pymoo library (Multi-objective Optimization in Python) \cite{blank2020pymoo}, which allows for easily using both single- and multi-objective algorithms, exploiting the default parameters proposed by the authors of the library.
Specifically, concerning CMAES, the initial standard deviation was set to 0.5 for each coordinate, and no restart strategy was applied.
Regarding DE, according to the classic DE taxonomy, the DE/rand/1/bin variant was used with a differential weight $F=0.3$ and a crossover rate equal to $0.5$.

\subsection{Standard benchmark functions}
\begin{figure*}[t]%
	\centering
	\subfloat[]{\includegraphics[width=\linewidth]{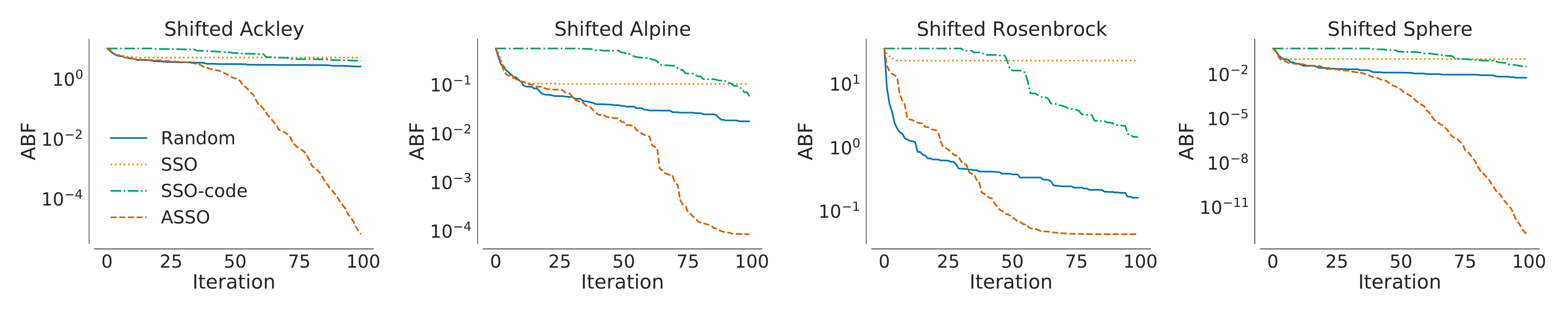}}\\%
	\subfloat[]{\includegraphics[width=\linewidth]{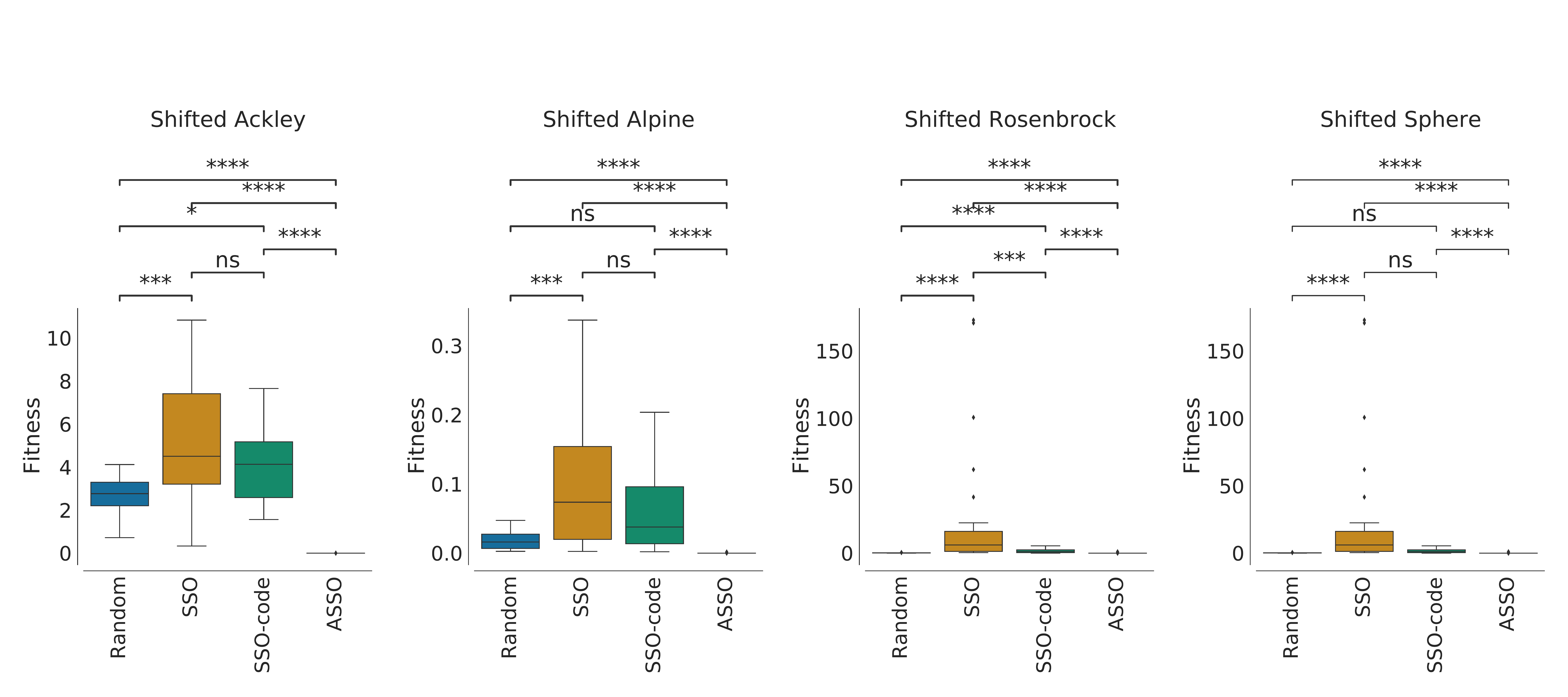}}%
	\caption
	{
		\textbf{(a)} Comparison of the ABF obtained by the analyzed techniques for each tested function.
		The ABF was calculated by using the fitness values of the best individual over the $30$ repetitions.
		To better appreciate the differences among the tested techniques, the $y$-axes are on a logarithmic scale.
		\textbf{(b)}The boxplots show the distribution of the last fitness value of the best individual over the $30$ repetitions. 
		p-value$\leq0.0001$ (****); $0.0001<$p-value$\leq0.001$ (***); $0.001<$p-value$\leq0.01$ (**); $0.01<$p-value$\leq0.05$ (*); p-value$>0.05$ (ns).
	}%
	\label{fig:figure1}%
\end{figure*}%
To show that both \ssa{} and \ssa{}-code are not shift-invariant, the search spaces of the tested benchmark functions were shifted by a large constant (i.e., $10^{9}$).
To collect statistically sound results, for each function, we ran the tested techniques $30$ times. 
For each iteration, we kept track of the fitness value of the best individual, over the $30$ repetitions, to calculate the \textit{Average Best Fitness} (ABF).
For a completely fair comparison among the different techniques, we fixed a budget of $100$ iterations using $50$ individuals.
Note that the implemented RS randomly generates $50$ particles, at each iteration, without taking into account any information of the previous iterations.

Figure \ref{fig:figure1}a clearly shows that both \ssa{} and \ssa{}-code are not shift-invariant.
Indeed, shifting $2$-$D$ standard benchmark functions by a large constant hampered the optimization abilities of both \ssa{} and \ssa{}-code.
Across all the tested functions, even RS obtained better results compared to \ssa{} and \ssa{}-code.
On the contrary, our proposed algorithm \asso{} -- where we simply removed the term $c_1 lb_j$ -- was able to outperform the other techniques.
It is worth reminding that \asso{} is not a novel algorithm, but an amended version of SSO in which the mathematical errors have been corrected.

In order to evaluate whether the achieved results were different also from a statistical point of view, we applied the Mann–Whitney U test with the Bonferroni correction \cite{mann1947,wilcoxon1992,dunn1961}.
Specifically, we applied this statistical test to independently compare the results obtained by the techniques on each benchmark function.
Thus, for each benchmark function and for each technique, we built a distribution by considering the fitness value of the best individual at the end of the last iteration over the $30$ repetitions.
The boxplots in Figure \ref{fig:figure1}b show the distribution of the best fitness values, achieved at the end of the executions of the tested techniques.
Figure \ref{fig:figure1}b also reports the results of the statistical tests by using the asterisk convention.
These results indicate that, generally, there is no statistical difference among \ssa{} and \ssa{}-code.
Only considering the Rosenbrock function a strong statistical difference is present between \ssa{} and \ssa{}-code.
The simple RS obtained similar or even better results than \ssa{} and \ssa{}-code, while there is always a strong statistical difference between the results achieved by \asso{} and those achieved by the other techniques, demonstrating the effectiveness of our correction.

\begin{figure*}[!t]
	\centering
	\includegraphics[width=.99\textwidth]{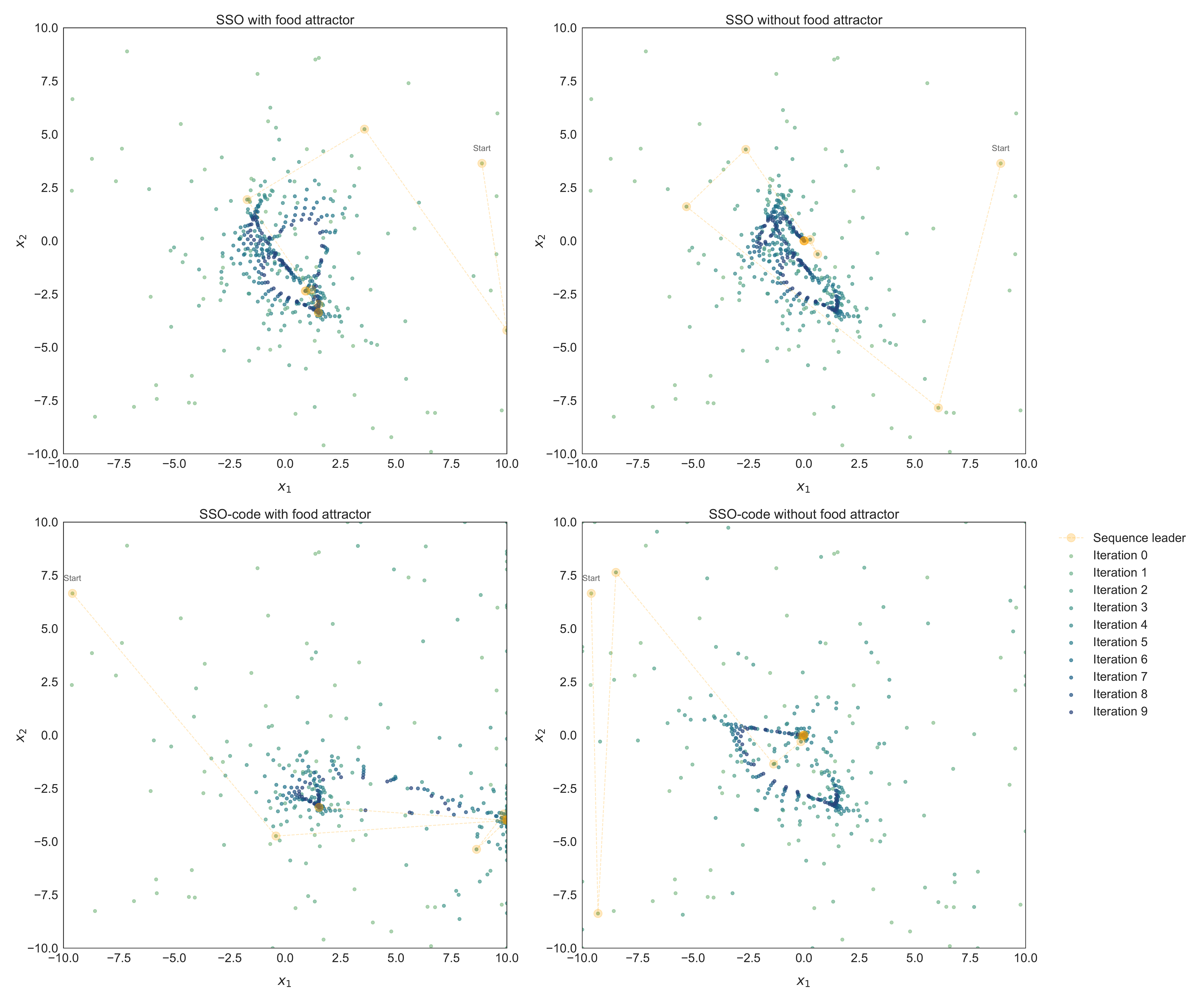}
	\caption{Dynamic behavior of the salp swarm: the leader leaves the feasible region, is relocated on the boundary, and then the swarm is attracted towards the origin of axis.}
	\label{fig:alldynamics}
\end{figure*}

Concerning the problem described in Section \ref{sec:update}, we show that, on a search space symmetric with respect to zero, the \ssa{} algorithm has a bias toward the origin.
In order to show this behavior, we will use a fitness function that returns a random number with uniform distribution in $[0,1)$.
For a swarm intelligence algorithm, we would expect a uniform distribution of particles across the entire search space using such a fitness function. 
Stated otherwise, using a random fitness function, the salps should not converge anywhere, and should randomly wander across the search space. 
However, we will show that, following the non-amended equations provided in the original \ssa{} paper \cite{MIRJALILI2017163}, the swarm converges in the origin providing an unfair advantage in the case of optimization problems whose global optimum lies in $\textbf{x}=\textbf{0}$ (and possibly lead to sub-optimal performance in the case of real-world functions).

Figure \ref{fig:alldynamics} shows the result of this test, performed on both SSO and SSO-code with and without the food attractor.
The Figure reports the positions of all salps during the first $10$ iterations of the optimization; the position of the leader salp is highlighted by an orange circle, and the initial position is denoted by the text ``Start''.
According to our results, the leader salp get attracted toward zero. The same happens for SSO-code: the leader salp it is inevitably attracted towards the center of the search space.
The attraction towards the origin is even more evident in the case of SSO and SSO-code without food attraction: the swarm perfectly converges to the origin and no longer moves.

\subsection{CEC 2017 benchmark function suite}

\begin{table}[!t]
	\centering
	\caption{Definitions and optimum values of the CEC 2017 benchmark functions designed for the competition on real-parameter single objective numerical optimization.}
	\label{table:benchmarks}
	\begin{scriptsize}
		\begin{tabular}{p{1.5cm}|p{0.35cm}|p{8cm}|p{0.5cm}}
			\hline\hline
			\textbf{Typology} & \textbf{No.} & \textbf{Function name} & \textbf{Opt.} \\ \hline 
			\multirow{3}{1.75cm}{\textit{Unimodal Functions}}
			& 1 & Shifted and Rotated Bent Cigar  & 100 \\
			& 2 & Shifted and Rotated Sum of Different Power  & 200 \\
			& 3 & Shifted and Rotated Zakharov  & 300 \\ \hline
			\multirow{7}{1.75cm}{\textit{Simple Multimodal Functions}}
			& 4 & Shifted and Rotated Rosenbrock & 400 \\
			& 5 & Shifted and Rotated Rastrigin  & 500 \\
			& 6 & Shifted and Rotated Expanded Schaffer F6  & 600 \\
			& 7 & Shifted and Rotated Lunacek Bi-Rastrigin  & 700 \\
			& 8 & Shifted and Rotated Non-Continuous Rastrigin  & 800 \\
			& 9 & Shifted and Rotated Levy  & 900 \\
			& 10 & Shifted and Rotated Schwefel  & 1000 \\ \hline
			\multirow{10}{1.75cm}{\textit{Hybrid Functions}}
			& 11 & Zakharov; Rosenbrock; Rastrigin & 1100 \\
			& 12 & High-conditioned Elliptic; Modified Schwefel; Bent Cigar & 1200 \\
			& 13 & Bent Cigar; Rosenbrock; Lunacek bi-Rastrigin & 1300 \\
			& 14 & High-conditioned Elliptic; Ackley; Schaffer F7; Rastrigin & 1400 \\
			& 15 & Bent Cigar; HGBat; Rastrigin; Rosenbrock & 1500 \\
			& 16 & Expanded Schaffer F6; HGBat; Rosenbrock; Modified Schwefel & 1600 \\
			& 17 & Katsuura; Ackley; Expanded Griewank plus Rosenbrock; Schwefel; Rastrigin & 1700 \\
			& 18 & High-conditioned Elliptic; Ackley; Rastrigin; HGBat; Discus & 1800 \\
			& 19 & Bent Cigar; Rastrigin; Griewank plus Rosenbrock; Weierstrass; Expanded Schaffer F6 & 1900 \\
			& 20 & HappyCat; Katsuura; Ackley; Rastrigin; Modified Schwefel; Schaffer F7 & 2000 \\ \hline
			\multirow{10}{1.75cm}{\textit{Composition Functions}}
			& 21 & Rosenbrock; High-conditioned Elliptic; Rastrigin & 2100 \\
			& 22 & Rastrigin; Griewank; Modified Schwefel & 2200 \\
			& 23 & Rosenbrock; Ackley; Modified Schwefel; Rastrigin & 2300 \\
			& 24 & Ackley; High-conditioned Elliptic; Griewank; Rastrigin & 2400 \\
			& 25 & Rastrigin; HappyCat; Ackley; Discus; Rosenbrock & 2500 \\
			& 26 & Expanded Schaffer F6; Modified Schwefel; Griewank; Rosenbrock; Rastrigin & 2600 \\
			& 27 & HGBat; Rastrigin; Modified Schwefel; Bent Cigar; High-conditioned Elliptic; Expanded Schaffer F6 & 2700 \\
			& 28 & Ackley; Griewank; Discus; Rosenbrock; HappyCat; Expanded Schaffer F6 & 2800 \\
			& 29 & $f_{15}$; $f_{16}$; $f_{17}$ & 2900 \\
			& 30 & $f_{15}$; $f_{18}$; $f_{19}$ & 3000 \\
			\hline\hline
		\end{tabular}
	\end{scriptsize}
\end{table}
During the last years, different benchmark function suites have been proposed to test and compare existing and novel global optimization techniques \cite{tangherloni2019ASoC}.
The benchmark functions they contain try to mimic the behavior of real-world problems, which often show complex features that basic optimization algorithms might not be able to grasp \cite{gallagher2016, tangherloni2019ASoC}.
Regarding the real-parameter numerical optimization, the IEEE Congress on Evolutionary Computation (CEC) and the Genetic and Evolutionary Computation Conference (GECCO) include competitions where complex benchmark function suites have been designed to test and compare global optimization techniques \cite{cec2017, bbob2018,hansen2016,doerr2018}.

\begin{figure*}[!t]
	\centering
	\includegraphics[width=.99\textwidth]{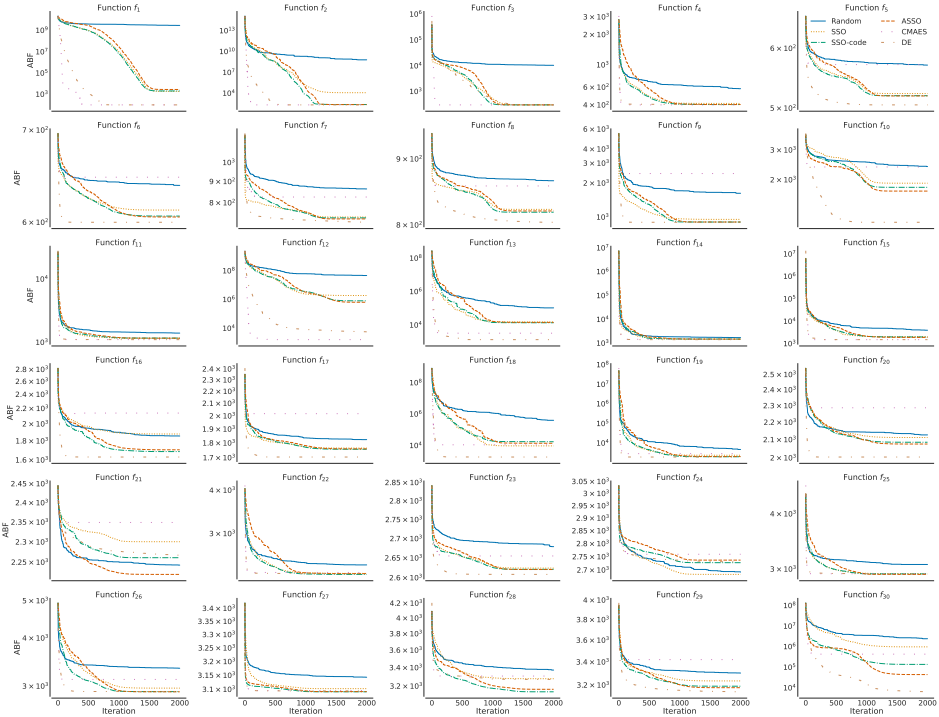}
	\caption{Comparison of the Average Best Fitness (ABF) obtained by the considered techniques for each function $f_{k}$ (with $k=1, \dots, 30$) and $D = 10$.
		The ABF was calculated by using the fitness values of the best individual over the $30$ repetitions. 
		Note that the y-axes are on a logarithmic scale.
	}
	\label{fig:ABFs10dims}
\end{figure*}

\begin{figure*}[!t]
	\centering
	\includegraphics[width=.99\textwidth]{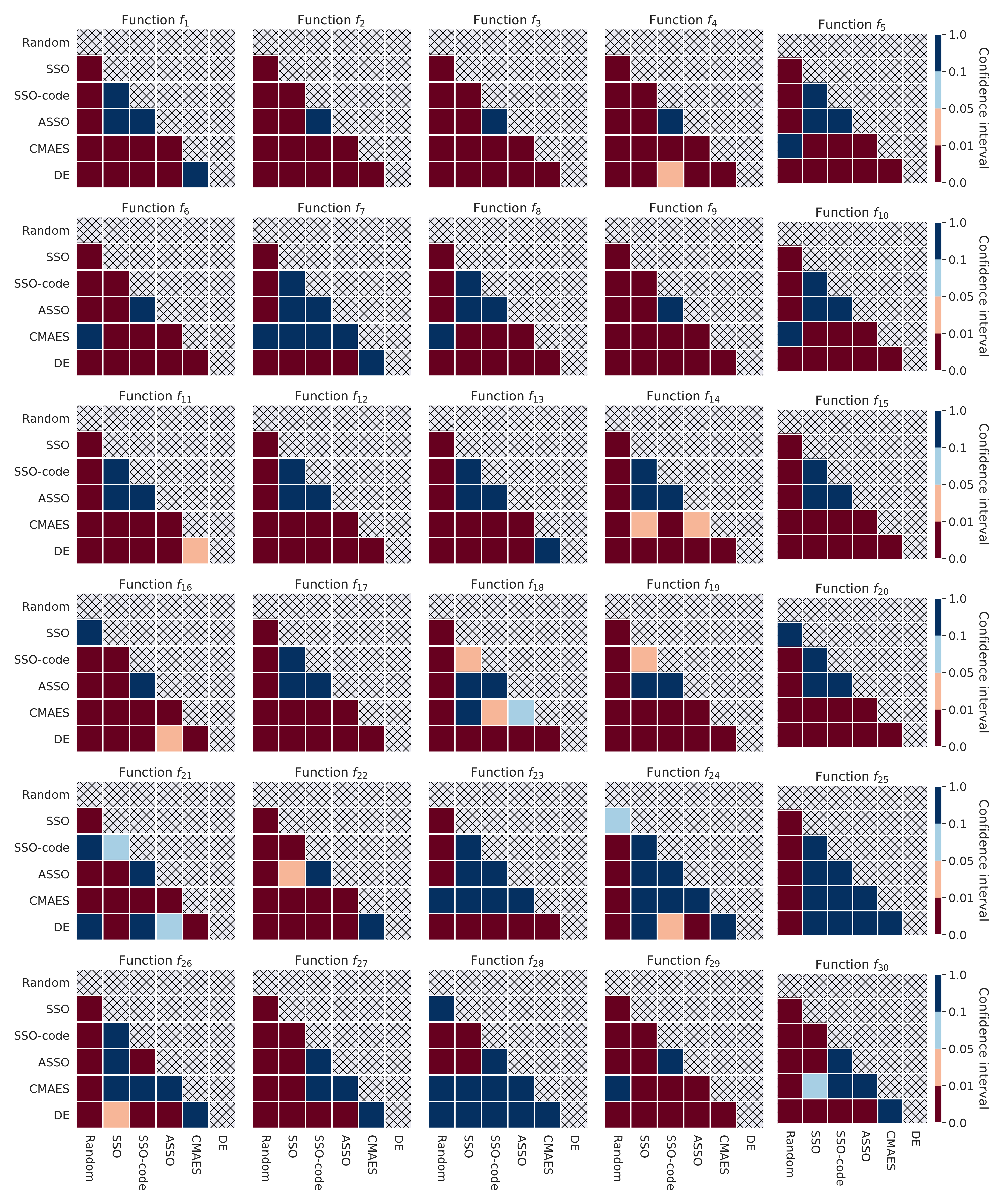}
	\caption{Heatmaps showing the p-values of the Mann–Whitney U test with the Bonferroni correction for each function $f_{k}$ (with $k=1, \dots, 30$) and and $D = 10$.
		The confidence interval was divided into $4$ levels indicating a strong statistical difference, statistical difference, weak statistical difference, and no statistical difference, respectively.}
	\label{fig:heatmaps10dims}
\end{figure*}

Here, we evaluated the performance of the considered optimization techniques using the CEC 17 benchmark problems for single-objective real-parameter numerical optimization \cite{awad2016problem}, which were previously used to compare the performance of different metaheuristics \cite{tangherloni2017,nobile2018cec,tangherloni2019ASoC}.
Table \ref{table:benchmarks} reports the tested benchmark problems, which are based on shifted, rotated, non-separable, highly ill-conditioned, and complex optimization benchmark functions \cite{awad2016problem}.
We optimized each function $f_{k}$ (with $k=1, \dots, 30$) considering the dimensions $D = \{10, 30\}$, search space boundaries $[-100, 100]^D$, and a maximum number of function evaluations $\texttt{MaxFES} = D \cdot 10^4$.
For each technique, we executed $30$ independent runs to collect statistically sound results.

Figure \ref{fig:ABFs10dims} depicts the Average Best Fitness (ABF) obtained by the analyzed techniques for each function $f_{k}$ (with $k=1, \dots, 30$) and $D = 10$, showing that DE was able to obtain better results than all the SSO-based strategies, including ASSO, and the simple RS tested here.
As one can see, DE outperformed the SSO-based strategies in 27 out of 30 functions.
CMAES also obtained better results than the SSO-based strategies in more than half of the tested functions.
As we did for the standard benchmark functions, we evaluated if the achieved results were also different from a statistical point of view by using the Mann–Whitney U test with the Bonferroni correction \cite{mann1947,wilcoxon1992,dunn1961}.
Thus, we independently compared the results obtained by the six techniques on each benchmark function $f_{k}$ (with $k=1, \dots, 30$) considering the distributions built using the value of the best individual at the end of the last iteration over the $30$ repetitions.
The heatmaps showed in Figure \ref{fig:heatmaps10dims} clearly point out that there is almost always a strong statistical difference between the results achieved by all the SSO-based strategies, including ASSO, and those obtained by DE.
Specifically, SSO-based strategies were able to obtain comparable or better results than those reached by DE only for the functions $f_{21}$, $f_{25}$, and $f_{28}$.
Comparing the results achieved by CMAES and those obtained by SSO-based strategies, there is a strong statistical difference in more than half of the tested functions.
These results are coherent with the probabilistic re-formulation of the  \emph{no-free lunch theorem} (NFL) \cite{lockett2017}, an extension of the original version of the NFL theorem \cite{wolpert1997no,macready1996makes}, which prove the validity of the theorem in continuous domains.
Thus, according to the NFL theorem, no algorithm outperforms all the competitors in any optimization problem.

\begin{figure*}[!t]
	\centering
	\includegraphics[width=.99\textwidth]{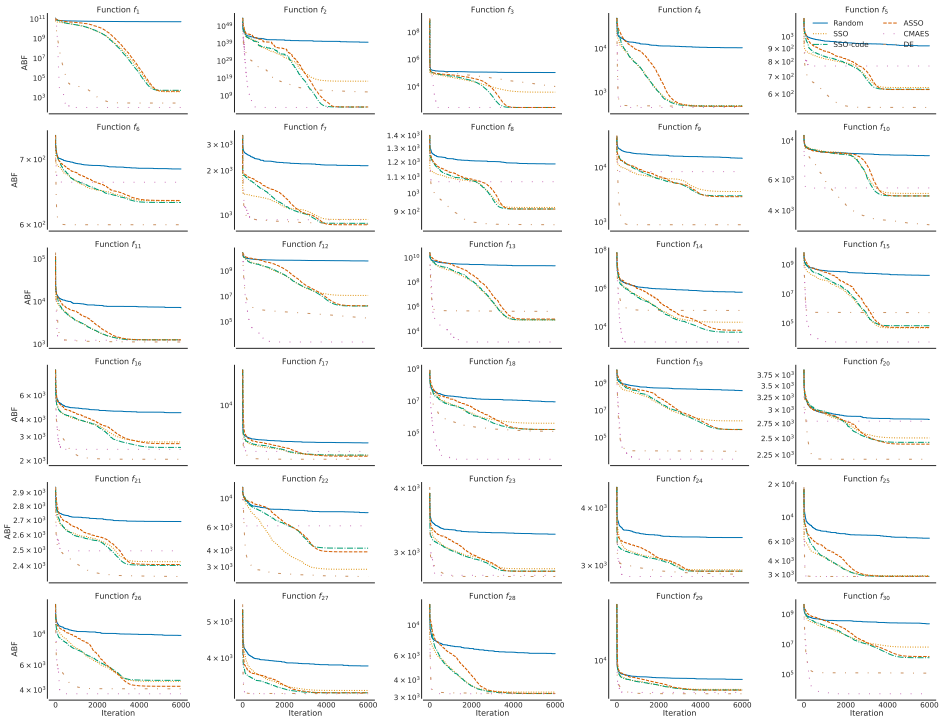}
	\caption{Comparison of the ABF obtained by the analyzed techniques for each function $f_{k}$ (with $k=1, \dots, 30$) and $D = 30$
		The ABF was calculated by using the fitness values of the best individual over the $30$ repetitions. 
		Note that the y-axes are on a logarithmic scale.
	}
	\label{fig:ABFs30dims}
\end{figure*}

\begin{figure*}[!t]
	\centering
	\includegraphics[width=.99\textwidth]{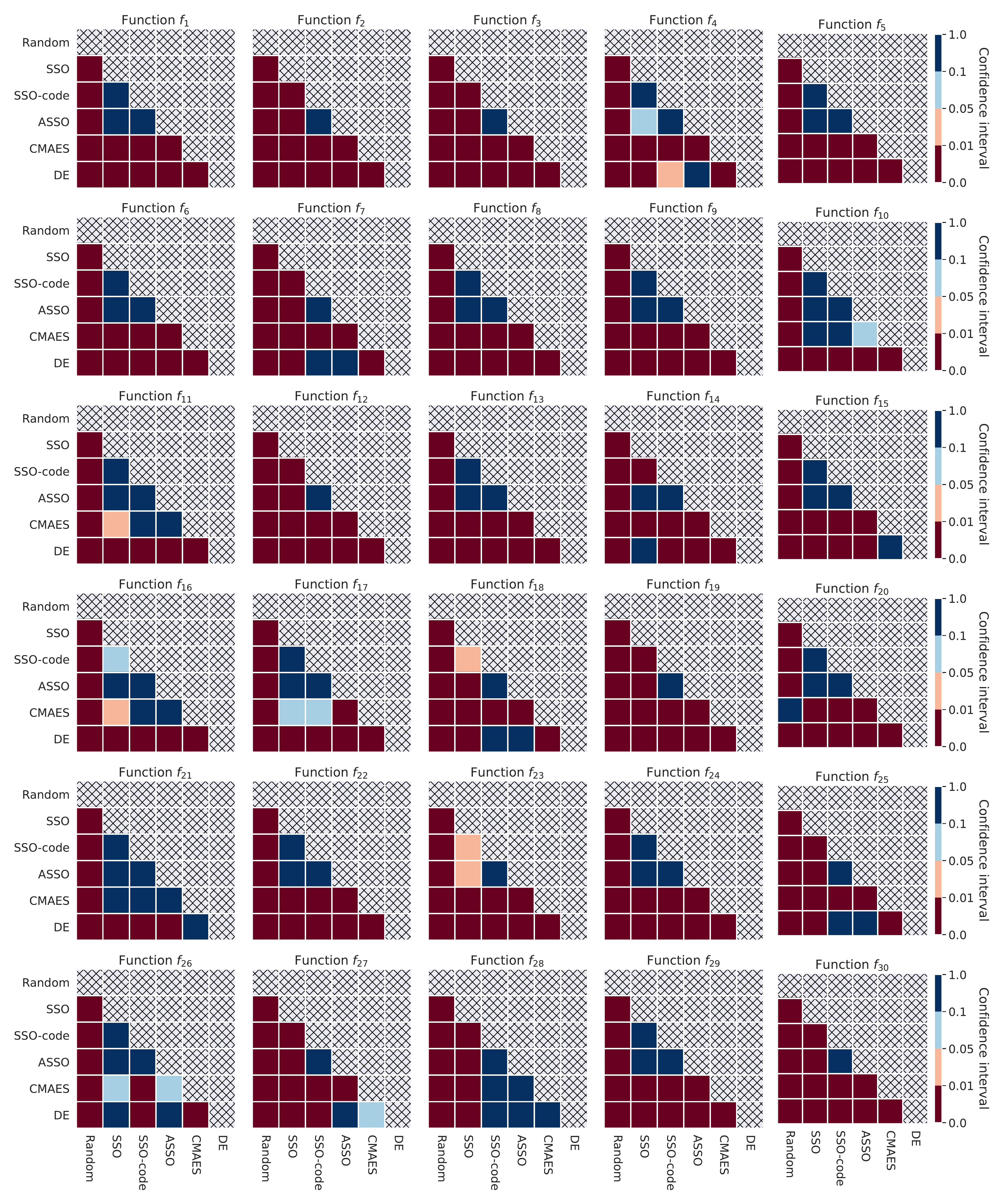}
	\caption{Heatmaps showing the p-values of the Mann–Whitney U test with the Bonferroni correction for each function $f_{k}$ (with $k=1, \dots, 30$) and and $D = 30$.
		The confidence interval was divided into $4$ levels indicating a strong statistical difference, statistical difference, weak statistical difference, and no statistical difference, respectively.}
	\label{fig:heatmaps30dims}
\end{figure*}

Increasing the dimensions of the tested benchmark functions from $10$ to $30$ does not allow the SSO-based approaches to obtain better results than DE.
As a matter of fact, Figures \ref{fig:ABFs30dims} and \ref{fig:heatmaps30dims} clearly show that DE outperformed all the SSO-based approaches in 22 out of 30 benchmarks functions.
In the remaining functions (i.e., $f_{4}$, $f_{7}$, $f_{14}$, $f_{18}$, $f_{25}$, $f_{26}$, $f_{27}$, and $f_{28}$) some SSO-based approaches (generally ASSO) obtained comparable results than those achieved by DE.
However, there is no benchmark function where all the SSO-based methods were able to obtain similar results to those reached by DE, confirming the poor performance of all the tested variants of SSO.

\section{Conclusion}
\label{sec:conc}

This paper shows systematic and deep issues in the definition of a widely-cited optimization algorithm, namely \extname{}. In particular, the multiple issues concerning the updating rules, the physically-inspired motivations, and the inconsistency between the code and the description in the original paper that proposed \extname{} raise concerns about all ensuing related literature, since the erroneous derivations and rules are present in most of the published papers on the topic. 
Furthermore, it is currently problematic to discern which results can be trusted, which ones are based on an incorrect implementation, and which papers have an incorrect description but are using a correct implementation.

The most serious issue analyzed in this paper is perhaps the presence of the lower bound $lb_j$ in the updating rule of the leader salp (see Equation~\ref{eq:3.1}).
In particular, this term makes the algorithm not shift-invariant, introducing a severe search bias that depends on the distance between the lower bound and the origin.
As shown in our experiments, under some specific circumstances, this factor can significantly affect the search capabilities of \ssa{}, which was outperformed by a simple random search. 
If the term $lb_j$ was inadvertently introduced, \emph{all} the current literature on \ssa{} contains results not reflecting the intended definition of the algorithm. 
On the other hand, if the authors meant to insert the term, the \ssa{} algorithm cannot work in spaces that have lower bounds too far from $0$ in all dimensions.
We also compared the described SSO-based approaches against DE and CMAES for the optimization of the CEC 2017 benchmark functions \cite{awad2016problem}, showing that all the SSO-based versions were outperformed by a simple DE version on almost all functions.
These results highlight, once more, that SSO and similar algorithms do not give any particular advantages with respect to widespread and common metaheuristics.
Considering all the results discussed in this work, we expect that more sophisticated metaheuristics, such as Linear population size reduction Successful History-based Adaptive DE (L-SHADE) \cite{tanabe2014improving, polakova2017shade, piotrowski2018step, piotrowski2018shade} that already showed their superior performance compared to DE, should outperform all the SSO-based approaches.
Thus, based on the evidences of this work, we discourage the use of SSO by the scientific community.
%There is no evidence supporting the use of SSO-based metaheuristics to address any optimization problem.
In particular, there is no theory supporting the convergence properties of SSO and its (supposed) superiority with respect to the existing metaheuristics.
On the contrary, SSO is defined and implemented based on a wrong mathematical formulation, as discussed in the first part of this paper.

We conclude this paper with a general remark about the metaphor-based approach for metaheuristics.
As mentioned in the Introduction, a lot of metaheuristic optimization algorithms have been proposed in the last years, most of them based on a particular natural process or animal behavior as a metaphor for the exploration and exploitation phases of the search space.
As noted by \cite{sorensen15}, one of the likely causes of this phenomenon is the excessive focus on the \emph{novelty} of such methods in part of the research community on metaheuristics.
This research approach, however, has the downside of shadowing the true search components of an optimization algorithm with terms and concepts borrowed from the considered metaphor. Therefore, it can happen that a ``novel'' metaheuristic optimization technique turns out to be just another well-known algorithm under a heavy disguise.
This is the case, for example, of three other recent Swarm Intelligence algorithms, namely: the Grey Wolf Optimizer, the Firefly Optimization Algorithm, and the Bat Algorithm, which were shown by \cite{dorigo2020grey} to have strong similarities with Particle Swarm Optimization.
While in this manuscript we showed and fixed the methodological and implementation flaws of \ssa{}, we believe that a closer inspection of the algorithm's underlying metaphor would also highlight a strong resemblance to other established swarm algorithms.

\section*{Availability}
All source code used for the tests is available on GitLab at the following address: \url{https://gitlab.com/andrea-tango/asso}.

\section*{Acknowledgment} This work was supported by national funds through the FCT (Funda\c{c}\~ao para a Ci\^encia e a Tecnologia) by the projects GADgET (DSAIPA/DS/0022/2018) and
the financial support from the Slovenian Research Agency (research core funding no. P5-0410).

\bibliographystyle{abbrv}
\bibliography{biblio}

\end{document}